\RequirePackage[l2tabu, orthodox]{nag}
\RequirePackage{snapshot}

\documentclass[9pt,twocolumn]{extarticle}

\makeatletter
\if@twocolumn
  \usepackage[dvips,letterpaper,top=0.5in, bottom=0.5in, left=0.75in, right=0.5in,includefoot,heightrounded]{geometry}
\else
  \usepackage[dvips,letterpaper,margin=1in,includefoot,heightrounded]{geometry}
\fi

\usepackage{srcltx}

\usepackage[russian,portuges,english]{babel}

\iflanguage{portuges}
    {\newcommand{\keywordname}{Palavras-chaves}}
    {\newcommand{\keywordname}{Keywords}}

\usepackage{amsmath}
\usepackage{amssymb,amsfonts}

\usepackage{abstract}

\usepackage{graphicx}
\usepackage[usenames,dvipsnames,svgnames,x11names]{xcolor}
\usepackage{subfigure}

\usepackage{booktabs}
\usepackage{numprint}

\usepackage{setspace}
\usepackage{flushend}

\usepackage{cite}

\usepackage{hyperref}
\usepackage[normalem]{ulem}

\usepackage{enumerate}

\usepackage{multirow}
\usepackage[noend]{algpseudocode}

\usepackage{listings}

\usepackage[short,12hr]{datetime}
       \usepackage{fouriernc}
\makeatletter

\newcommand{\printtitle}{%
\makeatletter
\if@twocolumn

\twocolumn[%
  \maketitle
  \begin{onecolabstract}
    \myabstract
  \end{onecolabstract}
  \begin{center}
    \small
    \textbf{\keywordname}
    \\\medskip
    \mykeywords
  \end{center}
  \bigskip
]
\saythanks
\else
  \maketitle
  \begin{onecolabstract}
    \myabstract
  \end{onecolabstract}
  \begin{center}
    \small
    \textbf{\keywordname}
    \\\medskip
    \mykeywords
  \end{center}
  \bigskip
  \onehalfspacing
\fi
\makeatother
}

\author{%
R.~J.~Cintra%
\thanks{R.~J.~Cintra
is
with
Signal Processing Group,
Departamento de Estat\'{\i}stica,
Universidade Federal de Pernambuco,
Recife, PE, Brazil.
At the time of writing,
R.~J.~Cintra
was
with
the LIRIS,
Institut National des Sciences Aplique\'es (INSA),
Lyon, France
(e-mail: rjdsc@de.ufpe.br).}
\and
Stefan Duffner%
\thanks{%
Stefan Duffner and Christophe Garcia
are with the
Universit\'e de Lyon,  CNRS, INSA-Lyon, LIRIS, UMR5205, F-69621, France.
(e-mail: \{\mbox{stefan.duffner},~christophe.garcia\}@liris.cnrs.fr}
\and
Christophe Garcia${}^\dagger$
\and
Andr\'e Leite%
\thanks{Andr\'e Leite
is with the
Departamento de Estat\'{\i}stica,
Universidade Federal de Pernambuco,
Recife, PE, Brazil.
(e-mail: leite@de.ufpe.br).}
}

\title{%
Low-complexity Approximate
Convolutional Neural Networks}

\newcommand{\myabstract}{%
In this paper,
we present an approach for minimizing the computational complexity
of trained Convolutional Neural Networks (\mbox{ConvNet}).
The idea is to approximate
all elements
of a given \mbox{ConvNet}
and
replace the original convolutional filters and parameters
(pooling and bias coefficients; and activation function)
with efficient approximations
capable of extreme reductions in computational complexity.
Low-complexity
convolution filters
are obtained through a
binary (zero-one)
linear programming scheme
based on the Frobenius norm
over
sets of dyadic rationals.
The resulting
matrices
allow for
multiplication-free computations
requiring
only addition and bit-shifting operations.
Such low-complexity structures
pave the way for
low-power, efficient
hardware designs.
We applied our approach on three use cases of different complexity:
(i)~a ``light'' but efficient \mbox{ConvNet} for face detection (with around \numprint{1000} parameters);
(ii)~another one for hand-written digit classification (with more than \numprint{180000} parameters);
and
(iii)~a significantly larger \mbox{ConvNet}: AlexNet with
$\approx$1.2~million matrices.
We evaluated the overall performance on the respective tasks for different levels of approximations.
In all considered applications,
very low-complexity approximations have been derived maintaining an almost equal classification performance.
}

\newcommand{\mykeywords}{%
Convolutional Neural Networks,
Approximation, Optimization,
Numerical Computation
}

\date{}

\begin{document}

\printtitle

\section{Introduction}

Since their introduction in the 1990s by LeCun~\emph{et~al.}~\cite{LecunBottouBengioHaffner},
convolutional neural networks (\mbox{ConvNet}s)
have proven to be very powerful in many challenging computer vision tasks, such as hand-written character recognition~\cite{LecunBottouBengioHaffner,ChellapillaPuriSimard06}, embedded text detection and recognition~\cite{DelakisG08,ElagouniGMS14}, automatic facial analysis~\cite{Vaillant94,CFF,OsadchyLeCunMillerPerona05}, traffic sign recognition~\cite{CiresanMMS12}, pedestrian detection~\cite{SermanetKCL13}, vision-based navigation~\cite{HadsellSermanetScoffier09},
and
house numbers recognition~\cite{SermanetCL12}, just to cite a few.
Although state-of-the-art results have been reached in many different fields, \mbox{ConvNet}s have become very popular only recently with the impressive results obtained by
Krizhevsky~\emph{et~al.}~\cite{krizhevskysh12}
in the recognition task, followed by Simonyan and Zisserman who won the localization challenge at the Large-Scale Visual Recognition Challenge (ImageNet) 2014~\cite{simonyanz14a}.

The main property of \mbox{ConvNet}s is
their capability for automatic extraction
of complex and application-suitable features from raw data (e.g., pixels in computer vision).
To do so, they integrate a pipeline of convolution and pooling layers generally followed by a multi-layer perceptron, jointly performing local feature extraction and classification (or regression) in a single architecture where all parameters are learnt using the classical error back-propagation algorithm~\cite{rumelhart86}.
Traditionally, \mbox{ConvNet}s are used for image processing tasks and, to this end, they are often applied on small regions of a bigger image in a sliding-window framework, for instance to detect an object.
Because of weight sharing, each layer essentially performs convolution or pooling operations using small kernels inside a ``retina'', and when applied to a large image, the replication of the \mbox{ConvNet} operation over all positions in the image can be significantly optimized by performing each convolution over the full image at once, efficiently implementing a full image transformation pipeline.

However, during training and when applying the trained \mbox{ConvNet}, there is still a significant number of floating-point computations (multiplications and additions).
In order to accelerate these operations, most current approaches and available software rely on parallel computing using GPUs~\cite{krizhevskysh12,SermanetEZMFL13} facilitated to some extent by the inherently parallel architectures of \mbox{ConvNet}s.
The recent trend in ``deep learning'' to use more and more complex models with millions of parameters requires enormous amounts of computational resources, in particular for training the models but also for applying the learnt \mbox{ConvNet}s.
Reducing the computational complexity of these models is thus of great interest to the research community as well as to industry.
Moreover,
such reduction in complexity of \mbox{ConvNet}s is necessary to implement them on devices with limited resources (such as mobile devices) in order to operate at acceptable speed, for example for real-time applications, and reduce the overall power consumption.

In this paper, we focus on drastically reducing the computational cost itself by proposing a post-training approximation scheme aiming at replacing all parameters of a \mbox{ConvNet} with low-complexity versions.
That is, for the convolution filters, only additions and bit-shifting operations
are performed---no multiplication is necessary.
Additionally,
the activation function is sought to be replaced with low-complexity alternatives.
Formulated as an optimization problem,
we adopt matrix approximation techniques
based on the Frobenius norm error
and
dyadic rational numbers represented
in terms of Canonical Signed Digit~(CSD) encoding.

Indeed,
a sound approximation theory is a necessary
step to facilitated hardware development.
This is illustrated in the case of image compression
where the most efficient coding schemes are
based on approximate matrices
realized in dedicated hardware~\cite{6575105,899842,doi:10.1080/00207217.2014.954634,coutinho2018pruned}.
We aim at introducing
an approach for approximating CNNs
in order to pave the way for future
efficient dedicated hardware design.

This work is organized as follows.
Section~\ref{section-related}
provides a literature review on the efficient
numerical implementation of neural networks and, in particular, \mbox{ConvNet}s.
Section~\ref{section-approximation}
details our approach
for approximating
the elements of a given \mbox{ConvNet}
aiming at designing low-complexity structures.
In
Section~\ref{section-experiments}
we present the results of our experimental evaluation of the proposed approach for two typical \mbox{ConvNet} architectures.
We assess the approximate \mbox{\mbox{ConvNet}s}
relative to their exact counterparts
in terms of
several figures of merit.
We conclude the paper in
Section~\ref{section-conclusion}.

\section{Related Work}
\label{section-related}

Although GPU implementations~\cite{scherer2010} allow for fast training and application of \mbox{ConvNet}s on sufficiently equipped platforms,
their integration on embedded systems for real-time applications may be more difficult
due to the limited amount of available resources on these devices,
which usually requires a good trade-off between performance and code size.
Several previous works have tackled this problem.
In early works~\cite{white1992, marchesi1993, kwan1993} weight parameters of neural networks have been represented as power-of-two integers.
Thus all multiplications can be operated as simple bit shifts.
Direct training of these networks was also possible by keeping a floating-point version of each weight parameter, or otherwise use a technique called ``weight dithering''~\cite{vincent1991}.
Simard and Graf~\cite{simard1994} extended this idea by encoding all other parameters as powers of two except the weights, i.e.\ neuron activations, gradients, and learning rates.
Later, Draghici~\cite{draghici2002} conducted a broader analysis on the computational power of neural networks with reduced precision weights.
Recently, Machado~\emph{et~al.}~\cite{machado2015} proposed a specific approximation scheme for sparse representation learning using only values of powers of two, and they integrated this quantization into the learning process.
Finally, in the work of Courbariaux~\emph{et~al.}~\cite{courbariaux2015}, all weights are encoded as binary values (1-bit), and for the training a floating-point version is still needed. Similarly, Kim and Paris~\cite{kim2015}, proposes a completely binary neural network. However, an initial real-valued training phase is required.

More recently, special attention has been paid to hardware implementations of \mbox{ConvNet}s, especially on FPGAs.
In~\cite{MamaletRG07}, for example, a high-level optimization methodology is applied to the implementation of the CFF face detector~\cite{CFF}.
They propose algorithmic optimizations and advanced memory management and transform the floating-point computation into fixed-point arithmetic.
Interestingly such coarse approximation could furnish very similar detection rates and low false-alarm rates on referenced datasets, for a roughly sevenfold gain of speed.
Later, this work has been extended in~\cite{FarrugiaMRYP09}, where the authors present for the first time several implementations of the CFF algorithm on FPGA, with a parallel architecture composed of a Processing Element ring and a FIFO memory, which constitutes a generic architecture capable of processing images of different sizes.
Farabet and LeCun~\cite{FarabetMATLC10} also propose a scalable hardware architecture  to  implement  large-scale \mbox{ConvNet}s, with a modular vision engine for large image processing, with FPGA and ASIC implementations.
Chakradhar~\emph{et~al.}~\cite{chakradhar2010} present a dynamically configurable FPGA co-processor that adapts to complex \mbox{ConvNet} architectures exploiting different types of parallelism.
A very low-complexity ASIC design of \mbox{ConvNet}s has been developed by
Chen~\emph{et~al.}~\cite{chen2014}, allowing for very high execution speeds and power consumption of state-of-the-art \mbox{ConvNet}s.
Finally, Zhang~\emph{et~al.}~\cite{zhang2015} propose a FPGA design strategy and algorithmic enhancements to optimize the computational throughput and memory bandwidth for any given \mbox{ConvNet} architecture.

Other recent works have focused on the \emph{algorithmic} and \emph{memory} optimizations of large-scale \mbox{ConvNet}s.
For example, Mamalet~\emph{et~al.}~\cite{MamaletG12} proposed different strategies for simplifying the convolutional filters (fusion of convolutional and pooling layers, 1D separable filters), in order to modify the hypothesis space, and to speed-up learning and processing times.
These convolutions can also effectively be performed by simple multiplications of the filters with the respective input images in the frequency domain~\cite{mathieu2014}. However, due to the overhead of the FFT, there is only a computational gain with larger filter sizes, and if a given filter can be reused consecutively for many input images.
Vanhoucke~\emph{et~al.}~\cite{vanhoucke2011} presented a set of different techniques to accelerate the computation of \mbox{ConvNet}s on CPU, mostly for Intel and AMD CPUs, exploiting for example SIMD instructions, memory locality, and fixed-point representations.
Also, many recent works~\cite{osawa2017evaluating,xue2013,sainath2013,denil2013,denton2014,yang2014,jaderberg2014,lebedev2015} have focused on reducing the complexity of convolution or fully-connected layers of large-scale \mbox{ConvNet}s by replacing the high-dimensional matrix or tensor multiplications with several low-rank matrix multiplications using different low-rank factorization methods, either at test-time or both for training and testing.
Although, large gains in computational and memory resources can be obtained on complex \mbox{ConvNet}s, these optimizations do not focus on hardware implementation and low-power constraints.

As opposed to many previous works that integrate the approximation process into
the learning~\cite{courbariaux2015binaryconnect,
DBLP:journals/corr/CourbariauxB16,
DBLP:journals/corr/RastegariORF16,
DBLP:journals/corr/CourbariauxBD14,
DBLP:journals/corr/MiyashitaLM16},
our approach operates on existing fully-trained models, that originally may have been aimed for standard PCs or more powerful architectures. Thus, our approximation scheme allows to integrate these models into hardware with much fewer resources.

\section{Approximation Approach}

\label{section-approximation}

\subsection{General Goal}

Our goal is to derive low-complexity
structures
capable of
reducing the computational costs
of a given \mbox{ConvNet}.
Ideally,
the following
two conditions are simultaneously
expected to be satisfied:
\begin{enumerate}[(i)]

\item
the computational elements of the \mbox{ConvNet}
(convolutional filters, sub-sampling coefficients, bias values,
and sigmoid function calls)
are replaced by corresponding low-complexity structures;

\item
the performance of the \mbox{ConvNet} is not significantly
degraded.

\end{enumerate}
However,
addressing both above conditions
proves to be a hard task.
In particular,
the large number of variables,
the non-linearities,
and
extremely long simulation times
prevents
such approach.
Also,
to the best of our knowledge,
literature
furnishes no mathematical result
linking the approximation of individual \mbox{ConvNet}
elements and the final \mbox{ConvNet} performance.
Thus,
we
adopt a greedy-like
heuristic
which
consists
in
individually simplifying each computational
structure of a \mbox{ConvNet}
in the hope of
finding
a resulting structure
capable of good performance~\cite{cormen2009introduction}.

In a \mbox{ConvNet}
two main types
of
mathematical elements are found:
(i)~matrix structures
and
(ii)~activation functions.
The matrix structures
are represented
by
convolution filter weights,
sub-sampling operations,
and
bias values;
whereas the activation function
is usually a non-linear function,
such as
the threshold,
piecewise-linear,
and
sigmoid functions~\cite{haykin1999neural}.

To approximate these two classes of elements,
different tools are required.
For the matrix-based
structures,
we selected
matrix approximation methods
as a venue
to derive
low-complexity computational elements~\cite{Cintra2011a,oliveira2015discrete,seber2007matrix,tablada2015class}.
For the activation,
we separate methods capable
of
approximating functions
with efficient digital implementation~\cite{tommisk2003efficient,zhang1996sigmoid,basterretxea2004approximation}.

\subsection{Low-complexity Matrix Structures}

In~\cite{tablada2015class,
Bouguezel2011,
Bouguezel2008,
Bouguezel2008a,
britanak2007discrete,
Cintra2011a},
several methods for deriving approximations
of discrete transform matrices---such as the
discrete cosine transform~\cite{ahmed1974discrete}---were proposed.
Let
$\mathbf{M}$ be an $N\times N$ given matrix.
For instance,
$\mathbf{M}$ can be a convolutional filter.
In this case,
a computational instantiation of $\mathbf{M}$
applied
to evaluate a single output pixel
requires in principle
$N^2$ floating point multiplications.
A typical \mbox{ConvNet} may contain
thousands
of convolutional filters.
For example,
the classical
architecture
described in~\cite{krizhevskysh12}
contains 244{,}760 filters,
which is nowadays
considered a relatively small network.
Therefore,
to minimize such a significant computational
cost,
we aim at
obtaining
a low-complexity
matrix
$\hat{\mathbf{M}}$
capable of
satisfying the following relation
in an optimal sense:
$
\mathbf{M}
\approx
\hat{\mathbf{M}}
.
$
The matrix
$\hat{\mathbf{M}}$
is said to be an approximation for~$\mathbf{M}$.
Such approximate convolutional filters
would allow the realization of computationally
intensive \mbox{ConvNet}s
in limited resources architectures.

In this paper,
a low-complexity matrix
is
a matrix of dyadic rational entries.
Dyadic rational numbers
are
fractions
of the form $m/2^n$,
where $n$ is a positive integer
and $m$ is an odd integer.
Such numbers are suitable for binary arithmetic.
Indeed,
a multiplication by a dyadic rational
consists of a multiplication by $m$
followed by a right shift of $n$ bits.
Because $m$ is an integer,
we can take full advantage of fixed-point arithmetic.
Indeed $m$ can be given a binary representation
with minimum number of adders,
aiming at multiplicative irreducibility.
Multiplicative irreducibility is attained whenever
the minimum number of additions to implement
a multiplication by $m$ is equal to the number of ones
in the binary representation of $m$~\cite{britanak2007discrete}.
Multiplicative irreducibility is often
obtained when the CSD representation
is considered~\cite{dempster1994constant}.
Therefore,
a multiplication by $m$
can be converted into
a sequence of additions and
bit-shifting operations.
As a consequence,
low-complexity matrices
are multiplierless,
 a very desirable property as floating-point operations
are much more costly than additions and bit-shifting operations.

Standard methods for matrix approximation
include:
inspection~\cite{cintra2014},
matrix parametrization~\cite{Bouguezel2011},
and
matrix factorization~\cite{Cintra2012}.
However,
since a typical \mbox{ConvNet} may contain from thousands to millions of
filters,
inspection-based approaches are not feasible.
Methods based on the parametrization
of the matrix elements are also ineffective
because
(i)~the elements
of convolutional filters
are usually not clearly
related,
i.e.,
they do not satisfy
identifiable mathematical relationships
and
(ii)~the elements are not repeated.
Additionally,
\mbox{ConvNet} filters
are not expected
to satisfy properties,
such as
symmetry
and
orthogonality,
which favors the derivation of approximations.
Thus,
methods based on matrix factorizations
are less adequate.

\subsection{Matrix Approximation by Linear Programming}

We
adopted a general approach to the problem
of obtaining
$\hat{\mathbf{M}}$
according to a optimization
problem
as described below:
\begin{align}
\label{equation-general-problem}
\hat{\mathbf{M}}
=
\arg
\min_\mathbf{T}
\operatorname{error}
\left(
\mathbf{M}
,
\mathbf{T}
\right)
.
\end{align}
The above optimization problem
can yield better approximate matrices
if an expansion factor $\alpha$
is introduced~\cite{britanak2007discrete}.
By adopting
the usual Frobenius norm~\cite{seber2007matrix}
as an error measure,
\eqref{equation-general-problem}
can be recast
according to the following
mixed
integer nonlinear programming (INLP) setup~\cite{lee2011mixed}:
\begin{align}
\label{equation-frobenius-problem}
(
\alpha^\ast,
\mathbf{T}^\ast
)
=
\arg
\min_{\alpha, \mathbf{T}}
\|
\mathbf{M}
-
\alpha
\cdot
\mathbf{T}
\|^2
,
\end{align}
where
$\alpha > 0$
is the real-valued expansion factor
and
$\| \cdot \|$
is the Frobenius norm~\cite{seber2007matrix}.
The choice of the Frobenius norm
is justified
by the following argument.
An approximate CNN
must have its elements
numerically `close' to
elements from the exact CNN.
Therefore,
a measure that takes into consideration
distance in a energy-based manner
(euclidean distance sense)
emerges naturally
as a means to guarantee
that the approximate
filtering structures
(e.g., convolution kernels)
are close to the exact counterpart.
The Frobenius norm satisfies the above rationale.
This analysis is confirmed
in~\cite{osawa2017evaluating}.

To ensure
that the candidate matrices~$\mathbf{T}$
have low complexity,
we limited
the
search space
of
the above problem
to
the matrices
whose elements
are defined over a
sets of dyadic rationals~$\mathcal{D}$.
Some particular sets are~\cite{tablada2015class,cintra2014}:
\begin{align*}
\mathcal{D}_1
=
&
\left\{
-1, 0, 1
\right\}
,
\\
\mathcal{D}_2
=
&
\left\{
-2, -1, 0, 1, 2
\right\}
,
\\
\mathcal{D}_3
=
&
\left\{
-4, -3, -2, -1, 0, 1, 2, 3, 4
\right\}
,
\\
\mathcal{D}_4
=
&
\left\{
-4, -3, -2, -1, -\frac{3}{4}, -\frac{1}{2}, -\frac{1}{4}, 0, \frac{1}{4}, \frac{1}{2}, \frac{3}{4}, 1, 2, 3, 4
\right\}
,
\\
\mathcal{D}_5
=
&
\left\{
-7, -6, -5, -4, -3, -2, -1, -\frac{3}{4},
-\frac{1}{2}, -\frac{1}{4}, 0,
\right.
\\
&
\quad
\left.
\frac{1}{4}, \frac{1}{2},
\frac{3}{4},
1, 2, 3, 4,5,6,7
\right\}
,
\\
\mathcal{D}_6
=
&
\left\{
 -4,      -\frac{15}{4},       -\frac{7}{2},      -\frac{13}{4},
\ldots,
\frac{13}{4},        \frac{7}{2},       \frac{15}{4},          4
\right\}
,
\\
\mathcal{D}_7
=
&
\left\{
-5,      -\frac{19}{4},       -\frac{9}{2},      -\frac{17}{4},
\ldots,
\frac{17}{4}, \frac{9}{2}, \frac{19}{4}, 5
\right\}
\\
\mathcal{D}_8
=
&
\left\{
-7,
-\frac{27}{4},
-\frac{13}{2},
-\frac{25}{4},
\ldots,
\frac{25}{4},
\frac{13}{2},
\frac{27}{4},
7
\right\}
.
\end{align*}
Sets
$\mathcal{D}_6$,
$\mathcal{D}_7$,
and
$\mathcal{D}_8$
possess uniformly spaced rationals.

A straightforward way of addressing~\eqref{equation-frobenius-problem}
is as follows.
Considering a given set of dyadic rationals
$\mathcal{D}$,
for each element of $\alpha \cdot \mathbf{T}$,
we simply
find
the closest neighbour of such element
in~$\mathbf{D}$.
Such approach can be efficiently implemented
by means of binary search.
However,
this approach is only effective
as long as~\eqref{equation-frobenius-problem}
remains
an unconstrained optimization problem.
Alternatively,
we can consider a more flexible approach
based on integer linear programming (ILP).
For fixed values of $\alpha$,
the mixed INLP problem posed in~\eqref{equation-frobenius-problem}
can be
efficiently solved by means of
binary (zero-one) linear programming.
In other words,
we aim at converting a nonlinear problem into a linear one.
Indeed,
let $m_{i,j}$, $i,j=1,2,\ldots,N$,
denote the entries
of~$\mathbf{M}$
and
$r \in \mathcal{D}$
be a dyadic rational.
We adopt
the following
binary decision variables:
\begin{align*}
x_{i,j}(r)
=
\begin{cases}
1, & \text{if $m_{i,j} = r$,}
\\
0, & \text{otherwise.}
\end{cases}
\end{align*}
For binary (zero-one) variables
we
have
$y^2 = y$,
where $y$ is a dummy variable.
This fact
paves the way
for the linearization of the above-mentioned optimization problem.
Therefore,
\eqref{equation-frobenius-problem}
can be re-written according to the following
binary
linear programming problem~\cite{papadimitriou1998combinatorial,bienstock2004integer,lenstra1979computational}:
\begin{align}
\label{equation-linear-programming-problem}
\min_{x_{i,j}(r)}
\enspace
\sum_{i=1}^N
\sum_{j=1}^N
\sum_{r \in \mathcal{D}}
\left(
r
-
\alpha
\cdot
m_{i,j}
\right)^2
\cdot
x_{i,j}(r)
,
\end{align}
subject to
\begin{align*}
\sum_{r\in\mathcal{D}}
x_{i,j}(r)
=
1
,
\qquad
i,j=1,2,\ldots,N
.
\end{align*}
The above constraint
is to ensure that each element
$m_{i,j}$
is approximated by a unique dyadic rational in $\mathcal{D}$.
The solution
of the above problem
is denoted
as
$x_{i,j}^{(\alpha)}$,
$i,j=1,2,\ldots,N$,
$r \in\mathcal{D}$,
being linked to the choice of $\alpha$.
Such binary (zero-one)
solution
can be employed to
compute the actual
entries
$t_{i,j}^{(\alpha)}$
of the low-complexity matrix
associated to the considered~$\alpha$
according to:
\begin{align}
t_{i,j}^{(\alpha)}
=
\sum_{r\in\mathcal{D}}
r
\cdot
x_{i,j}^{(\alpha)}(r)
.
\end{align}
The resulting low-complexity
matrix
is denoted by
$\mathbf{T}_\alpha$.
The approximation error
is implied
by \eqref{equation-frobenius-problem}
and
can be computed
according
to:
\begin{align*}
\operatorname{Error}
(\alpha)
=
\|
\mathbf{M}
-
\alpha
\cdot
\mathbf{T}_\alpha
\|^2
.
\end{align*}

Because
a sequence of values for $\alpha$ is selected,
the above problem is solved for each instantiation;
furnishing
the sequence
of errors indexed by~$\alpha$:~$\operatorname{Error}(\alpha)$.
Being a linear programming problem,
each instantiation
can be solved efficiently and very quickly
by contemporary computational packages~\cite{matlab2010version,rct2013r:}.
State-of-the-art solvers
can obtain solutions for
ILP problems
at an average computation complexity
in
$\mathcal{O}(N)$~\cite{megiddo1984linear}
or
$\mathcal{O}(N \log N)$~\cite{lenstra1979computational,papadimitriou1998combinatorial,bienstock2004integer}.
Finally,
we determine the global optimum value~$\alpha^\ast$
according to:
\begin{align}
\alpha^\ast
=
\arg
\min_\alpha
\operatorname{Error}(\alpha)
,
\end{align}
which can be solved by simple minimization
over a vector of values.
Associated to $\alpha^\ast$,
we also obtain
$\mathbf{T}^\ast \triangleq \mathbf{T}_{\alpha^\ast}$,
which is the global optimal low-complexity
matrix.
Therefore,
the sought
approximation~$\hat{\mathbf{M}}$
is given by:
\begin{align}
\label{equation-approximation}
\hat{\mathbf{M}}
=
\alpha^\ast
\cdot
\mathbf{T}^\ast
.
\end{align}

The above ILP approach
allows the user to easily
include constraints to the optimization problem.
This is relevant
for further investigation in this topic;
in particular when
specific mathematical properties
are expected to be enforced on
the resulting
low-complexity matrices
(for instance,
2D~filter normalization~\cite[p.~115]{burger2016digital}).

We emphasize that
the solving method for~\eqref{equation-frobenius-problem}
is only required to be efficient enough
to cope with the time constraints
at the design phase of the
approximate neural network.
In other words,
solvers available in contemporary optimization packages
are suitable;
and
the choice of the particular method
for solving~\eqref{equation-frobenius-problem}
is not a critical for our approach.
Additionally,
we note that
the optimization solver is simply
a step for obtaining the final neural network.
Once the approximate structures are found,
optimization solver are
clearly
not required anymore.

\subsection{Example}

To illustrate
the procedure,
we selected
$
\mathcal{D}_8$
and
considered
the search space
of
the expansion factor
to be the interval $[0.25, 1]$
with a step of $10^{-3}$.
Additionally,
we consider the following particular
convolutional filter employed
in the \mbox{ConvNet} described in Section~\ref{sec:mnist}:
\begin{align*}
\mathbf{M}_0 =
\left[
\begin{smallmatrix}
1.5200701 &
1.0317051 &
0.7906240 &
-0.2153791 &
-0.2340538
\\
\\
1.3982610 &
2.1860176 &
2.0152923 &
1.5620477 &
0.8270900
\\
\\
-0.6848867 &
0.7470516 &
1.6923728 &
1.2537112 &
1.1946758
\\
\\
-1.2387477 &
-0.5483563 &
0.1261987 &
0.8677799 &
0.7742613
\\
\\
-1.4691808 &
-1.2178997 &
-0.2924347 &
0.2172496 &
0.1325074
\end{smallmatrix}
\right]
.
\end{align*}
Solving~\eqref{equation-frobenius-problem}
for the above matrix,
we obtain:
\begin{align*}
\alpha^\ast
&
=
0.30931
,
\\
\mathbf{T}^\ast
&
=
\begin{bmatrix}
 5 & 3.25 & 2.5 & -0.75 & -0.75 \\
 4.5 & 7 & 6.5 & 5 & 2.75 \\
 -2.25 & 2.5 & 5.5 & 4 & 3.75 \\
 -4 & -1.75 & 0.5 & 2.75 & 2.5 \\
 -4.75 & -4 & -1 & 0.75 & 0.5 \\
\end{bmatrix}
\\
&
=
\frac{1}{4}
\cdot
\begin{bmatrix}
   20 &  13 &  10 &  -3 &  -3 \\
   18 &  28 &  26 &  20 &  11 \\
   -9 &  10 &  22 &  16 &  15 \\
  -16 &  -7 &   2 &  11 &  10 \\
  -19 & -16 &  -4 &   3 &   2 \\
\end{bmatrix}
.
\end{align*}

\begin{figure*}
\centering
\subfigure[]{\includegraphics{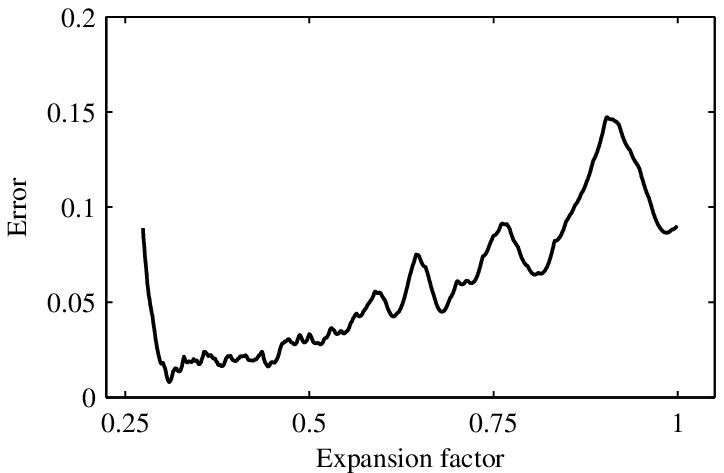}}
\subfigure[]{\includegraphics{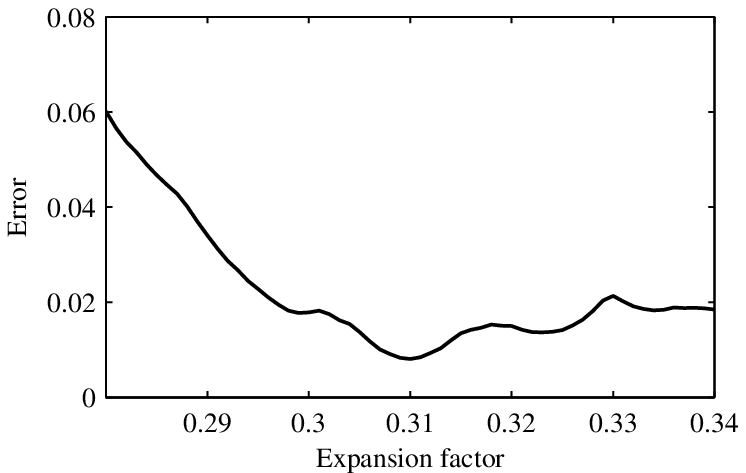}}
\caption{Approximation error for the particular matrix~$\mathbf{M}_0$:
(a)~error curve over the considered search interval
for $\alpha$
and
(b)~detailed view around the
optimum value of $\alpha$.
}
\label{figure-quadratic}

\end{figure*}

Fig.~\ref{figure-quadratic}(a) depicts
the Frobenius norm error
for varying  values of $\alpha$
(cf.~\ref{equation-frobenius-problem}).
For very small $\alpha$,
the values of $\alpha\cdot\mathbf{M}$ are close to zero.
So the discrete entries of the candidate matrices~$\mathbf{T}$
are unable to provide a good approximation.
As $\alpha$ increases,
a similar effect happens.
However,
for intermediate values the minimum can be found.
Fig.~\ref{figure-quadratic}(b) shows details in the
vicinity of the optimum.
The curves shown in Fig.~\ref{figure-quadratic}
are piecewise concatenations of parabolae.
This is due to the quadratic nature of the
coefficients
$(r - \alpha \cdot m_{i,j})^2$
of the
linear programming problem in~\eqref{equation-linear-programming-problem}.
Each parabola
is linked to a particular approximate candidate~$\mathbf{T}$.

Notice that the low-complexity matrix~$\mathbf{T}^\ast$
is expressed in terms of small integers,
which can be given simple binary expansions
(e.g., $22= 2^6-2^4-2^2$).
Similarly,
we have that
$
\alpha^\ast
=
0.30931
\approx
2^{-2}
+
2^{-4}
-
2^{-8}
=
0.30859375
$.

Therefore,
considering~\eqref{equation-approximation},
the actual fully multiplierless
approximation is
furnished
by:
\begin{align*}
\hat{\mathbf{M}}
=
(
2^{-4}
+
2^{-6}
-
2^{-10}
)
\cdot
\begin{bmatrix}
   20 &  13 &  10 &  -3 &  -3 \\
   18 &  28 &  26 &  20 &  11 \\
   -9 &  10 &  22 &  16 &  15 \\
  -16 &  -7 &   2 &  11 &  10 \\
  -19 & -16 &  -4 &   3 &   2 \\
\end{bmatrix}
.
\end{align*}

\subsection{Activation Function Approximations}

Although there are several types
of activation functions,
we focus our analyses
on the
continuous tanh-sigmoid function,
which is defined
according to the hyperbolic tangent function~\cite{haykin1999neural}.
As indicated in~\cite{lecun1989generalization,haykin1999neural},
the mathematical expression for the tanh-sigmoid functon
is given by:
\begin{align}
\label{equation-tanh-sigmoidal}
\phi(x) = a \cdot \tanh(b \cdot x)
,
\end{align}
where
$a= 1.7159$
and
$b = 2/3$.
This particular activation function has been originally proposed by LeCun~\cite{lecun1988} and
adopted in several working models
as the
Convolutional Face Finder (CFF)~\cite{CFF}.

In~\cite{zhang1996sigmoida,tommisk2003efficient,basterretxea2004approximation,larkin2006,temam2012,schlessman2002approximation},
approximations
for the related sigmoid function
given by $y = 1/(1+e^{-x})$ were examined,
including
the Alippi and Storti-Gajani (ASG) approximation~\cite{alippi1991simple},
the piecewise linear approximation of a non-linear function (PLAN)~\cite{amin1997piecewise},
and
simple linear~\cite{tommisk2003efficient}
and quadratic~\cite{zhang1996sigmoida} approximations.
Based on these approximations,
we derived expressions for
the tanh-sigmoid approximations
as shown in Table~\ref{table-tanh-sigmoid-approximations}.
We adopted an 8-bit representation for
$a$ resulting in the
following approximate value:
$\hat{a} = 7/4$.

\begin{table*}
\centering
\caption{Approximations for the tanh-sigmoid
}

\label{table-tanh-sigmoid-approximations}

\begin{tabular}{lr|lr}
\toprule
ASG-based
&
$
\sigma_1(x)=
\hat{a} \cdot
\begin{cases}
-1 + \frac{1+ \frac{|\lfloor x \rfloor| - x}{2}}{2^{|x|}}, & \text{$x<0$,} \\
1 - \frac{1+ \frac{|\lfloor x \rfloor| - x}{2}}{2^{|x|}}, & \text{$x\geq 0$.}
\end{cases}
$
&
PLAN-based
&
$
\sigma_2(x)
=
\hat{a} \cdot
\begin{cases}
-1, & \text{$x < -5$,} \\
\frac{x}{16} - \frac{89}{128}, & \text{$-5\leq x<-\frac{19}{8}$,} \\
\frac{x}{4} - \frac{1}{4}, & \text{$-\frac{19}{8}  \leq x< -1$,} \\
\frac{x}{2}, & \text{$-1 \leq x< 1$,} \\
\frac{x}{4} + \frac{1}{4}, & \text{$1 \leq  x< \frac{19}{8}$,} \\
\frac{x}{16} + \frac{11}{16}, & \text{$\frac{19}{8}\leq x<5$,} \\
1, & \text{$x \geq 5$.}
\end{cases}
$
\\
\midrule
Linear I
&
$
\sigma_3(x)
=
\hat{a} \cdot
\begin{cases}
-1, & \text{$x < -4$,} \\
\frac{x}{4}, & \text{$-4 \leq x < 4$,} \\
1, & \text{$x \geq 4$.}
\end{cases}
$
&
Linear II
&
$
\sigma_4(x)
=
\hat{a} \cdot
\begin{cases}
-1, & \text{$x < -2$,} \\
\frac{x}{2}, & \text{$-2 \leq x < 2$,} \\
1, & \text{$x \geq 2$.}
\end{cases}
$
\\
\midrule
Quadratic I
&
$
\sigma_6(x)
=
\hat{a} \cdot
\begin{cases}
-1, & \text{$x < -4$,} \\
(\frac{x}{4}+1)^2 -1, & \text{$-4 \leq x <0$,} \\
1 - (\frac{x}{4}+1)^2, & \text{$0 \leq x <4$,} \\
1, & \text{$x \geq4$.}
\end{cases}
$
&
Quadratic II
&
$
\sigma_6(x)
=
\hat{a} \cdot
\begin{cases}
-1, & \text{$x < -2$,} \\
(\frac{x}{2}+1)^2 -1, & \text{$-2 \leq x <0$,} \\
1 - (\frac{x}{2}+1)^2, & \text{$0 \leq x <2$,} \\
1, & \text{$x \geq2$.}
\end{cases}
$
\\
\bottomrule
\end{tabular}

\end{table*}

\subsection{Complexity}

As a consequence of the above approximations,
we have substantial savings in computation costs.
Indeed,
a single call
of the original $N\times N$ matrix~$\mathbf{M}$
requires $N^2$ multiplications
of floating-point entries
per pixel.
On the other hand,
the proposed approximation~$\hat{\mathbf{M}}$
contains only small integers
that can be very efficiently encoded
with minimal number of adders~\cite{dempster1994constant}.
Similarly,
the expansion factor $\alpha^\ast$
can be given a truncated rational approximation
in the form of dyadic rationals.
The same rationale also applies
to
the remaining computational structures of the original \mbox{ConvNet}.
Thus,
the final resulting structure
is fully \emph{multiplierless}---only
additions and bit-shifting operations
are required.
In terms of hardware realization,
the number of arithmetic operations
translate
into chip area and power consumption~\cite{blahut2010fast,potluri2014improved}.
Thus,
in limited resource scenarios
(e.g., embedded systems and wireless sensors),
approximations
may
provide an effective way
of
porting large \mbox{ConvNet}s
into physical realization.

To summarize,
the proposed approximation approach consists of:
\begin{enumerate}[(i)]

\item
finding approximate convolutional filter
by solving~\eqref{equation-general-problem}
for each exact convolutional filter from a given \mbox{ConvNet};

\item
converting
scaling factors,
sub-sampling coefficients,
and bias values
into CSD representation
aiming at the minimization
of computation costs
and
multiplicative irreducibility;

\item
approximating
the
activation function to
a simple function.

\end{enumerate}

\section{Experiments}
\label{section-experiments}

We studied the effectiveness of the proposed approximation approach on two classical computer vision problems:
(i)~a binary
and
(ii)~a multi-class classification problem.
The first application is face detection, where the \mbox{ConvNet} classifies image regions as face or non-face.
The second one is handwritten digit recognition, where the trained model is used to classify a given image patch into one of the ten digits ``0'' to ``9''.
For each of the two applications, we trained a \mbox{ConvNet} in a classical way
and evaluated its performance in terms of precison and recall,
for the given application.

The first \mbox{ConvNet} is relatively small, whereas the second model (for digit recognition) contains much more parameters.
We aim at demonstrating
that our proposed approach is able to effectively process larger networks.

After approximating the parameters of the models, we compared their performance
with
their respective original, exact versions.
Note that we do not aim at improving the state-of-the-art in face detection or hand-written digit recognition.
Indeed,
current literature presents concrete example of complex models,
such as
multi-view or part-based detectors for face detection~\cite{mathias2014} and
huge ensemble classifiers for digit recognition~\cite{ciresan2012}.
Our goal is to demonstrate---based on common representative models---that the complexity of a given trained \mbox{ConvNet} model
can be reduced significantly by approximating its parameters while maintaining a very similar performance.

Hereafter
an approximate network
based on dyadic set $\mathcal{D}_i$
is referred to as
$A_i$.

\subsection{Binary Classification}

Our first set of experiments employs a \mbox{ConvNet}
that was trained for face detection in grey-scale images.
Thus, such network is a binary classifier that decides whether the given input image is a face or not.
As a working model,
we selected the
classical face detector called Convolutional Face Finder (CFF) proposed by Garcia and Delakis~\cite{CFF}.
This model is a relatively ``light'' \mbox{ConvNet} with
an input size of $32\times36$ and six layers:
four~layers alternating convolution and average pooling operations,
with
4, 4, 14, and 14~maps, respectively,
followed by 14~neurons and one single final output neuron.
The first convolution layer contains four filters of size $5\times5$, the second one contains
20~filters of size~$3\times3$,
and
the
14~neurons of the first neuron layer
are treated as convolutions of size~$6\times7$,
each neuron being connected to only one map.
Pooling maps contain a single coefficient, and all maps and neurons have an additional bias.
The entire \mbox{ConvNet} has 951~trainable parameters in total.
A thorough description of this particular
\mbox{ConvNet} is supplied in~\cite{CFF}.
The employed activation function is
the exact continuous tanh-sigmoidal function as detailed in~\eqref{equation-tanh-sigmoidal}.

After training the \mbox{ConvNet} as described in~\cite{CFF}, we approximated all the convolution filter matrices with low-complexity versions. %
We created several approximations using the different sets of dyadic rationals described in the previous section:
$\mathcal{D}_1, \mathcal{D}_1, \ldots, \mathcal{D}_8$.
We also replaced all average pooling coefficients and bias terms with their closest CSD representation using 8~bits,
being 7~bits for the fractional part.

Table~\ref{table-cost-cff}
lists the arithmetic costs
of the exact \mbox{ConvNet}
compared to its approximations.
Floating-point multiplications,
direct additions,
additions due to the CSD expansion,
and bit-shifting operations
were counted.
The exact structure requires
both floating-point multiplications and additions.
In contrast,
the approximate methods
completely eliminates the need for multiplications
at the expense
of much simpler operations:
additions and bit-shifting operations.
Because the approximate quantities can be easily represented in
fixed-point arithmetic representation,
it is suitable for hardware implementation.
Additionally,
the hardware implementation of bit-shifting operations
require virtually no cost,
because it can be implemented by simple physical wiring.
As a result,
we have a very favorable
trade-off:
multiplications are exchanged for additions.

\begin{table*}
\centering
\caption{Arithmetic cost for CFF-based models}
\label{table-cost-cff}

\begin{tabular}{lcccc}
\toprule
\multirow{2}{*}{Model}
&
\multicolumn{4}{c}{Operation}
\\
\cmidrule{2-5}
&
Mult. & Add. & CSD Add. & Bit-shifting \\
\midrule
Exact &
882 & 843 & - & -
\\
$A_1$ &
0 & 843 & 235  &  346
\\
$A_2$ &
0 & 843 & 251  &  362
\\
$A_3$ &
0 & 843 & 377  &  488
\\
$A_4$ &
0 & 843 & 457  &  568
\\
$A_5$ &
0 & 843 & 506  &  617
\\
$A_6$ &
0 & 843 & 756  &  867
\\
$A_7$ &
0 & 843 & 842  &  953
\\
$A_8$ &
0 & 843 & 1028  &  1139
\\
\bottomrule
\end{tabular}
\end{table*}

In order to analyse the effect of the approximation on the actual performance of the \mbox{ConvNet},
we evaluated the different
versions on three standard face detection benchmarks:
FDDB~\cite{fddbTech} (\numprint{2845} images),
AFW~\cite{zhu2012} (205 images),
and
Pascal Faces dataset~\cite{yan2013} (851~images);
and we used the improved annotation and evaluation protocol proposed by Mathias~\emph{et~al.}~\cite{mathias2014}.

Tables~\ref{tab:cff_fddb}-\ref{tab:cff_pascal} show the average precision rates for different combinations of sigmoid and weight matrix approximations relative to the exact model for the three datasets.

\begin{table*}
\centering

\caption{%
Average precision for CFF with the FDDB test set and different approximations relative to the exact model}
\label{tab:cff_fddb}

\begin{tabular}{c|ccccccc}
\toprule
&
Exact & ASG & PLAN & Linear I & Linear II & Quadratic I & Quadratic II
\\
\midrule
Exact & 1.000 & 0.953 & 0.894 & 0.002 & 0.988 & 0.887 & 0.938 \\
$A_1$ & 0.000 & 0.000 & 0.000 & 0.000 & 0.000 & 0.000 & 0.000 \\
$A_2$ & 0.001 & 0.001 & 0.000 & 0.000 & 0.001 & 0.000 & 0.002 \\
$A_3$ & 0.549 & 0.311 & 0.432 & 0.005 & 0.510 & 0.426 & 0.180 \\
$A_4$ & 0.523 & 0.602 & 0.365 & 0.001 & 0.558 & 0.383 & 0.602 \\
$A_5$ & 0.490 & 0.098 & 0.517 & 0.000 & 0.657 & 0.510 & 0.073 \\
$A_6$ & 0.938 & 0.914 & 0.817 & 0.002 & 0.888 & 0.797 & 0.949 \\
$A_7$ & 0.960 & 0.943 & 0.847 & 0.002 & \textbf{0.963} & 0.848 & 0.935 \\
$A_8$ & 0.821 & 0.792 & 0.648 & 0.001 & 0.810 & 0.650 & 0.793 \\
$A_{7,3,3,3}$ & 0.917 & 0.902 & 0.886 & 0.003 & 0.947 & 0.888 & 0.851 \\
$A_{7,4,4,4}$ & 0.967 & 0.931 & 0.855 & 0.000 & \textbf{0.976} & 0.861 & 0.928 \\
$A_{7,6,6,6}$ & 0.959 & 0.907 & 0.862 & 0.004 & \textbf{0.959} & 0.863 & 0.933 \\
\bottomrule
\end{tabular}

\end{table*}

\begin{table*}
\centering

\caption{Average precision for CFF with the AFW test set and different approximations relative to the exact model}
\label{tab:cff_afw}

\begin{tabular}{c|ccccccc}
\toprule
&
Exact & ASG & PLAN & Linear I & Linear II & Quadratic I & Quadratic II
\\
\midrule
Exact & 1.000 & 0.839 & 0.829 & 0.000 & 1.041 & 0.797 & 0.383 \\
$A_1$ & 0.000 & 0.000 & 0.000 & 0.000 & 0.000 & 0.000 & 0.000 \\
$A_2$ & 0.001 & 0.001 & 0.000 & 0.000 & 0.000 & 0.000 & 0.000 \\
$A_3$ & 0.220 & 0.041 & 0.260 & 0.004 & 0.199 & 0.248 & 0.014 \\
$A_4$ & 0.422 & 0.356 & 0.317 & 0.000 & 0.457 & 0.336 & 0.275 \\
$A_5$ & 0.220 & 0.015 & 0.314 & 0.000 & 0.217 & 0.251 & 0.002 \\
$A_6$ & 0.864 & 0.794 & 0.715 & 0.000 & 0.893 & 0.694 & 0.700 \\
$A_7$ & 0.978 & 0.844 & 0.753 & 0.000 & \textbf{0.985} & 0.755 & 0.614 \\
$A_8$ & 0.698 & 0.553 & 0.576 & 0.000 & 0.565 & 0.544 & 0.169 \\
$A_{7,3,3,3}$ & 0.955 & 0.525 & 0.835 & 0.004 & 0.914 & 0.816 & 0.184 \\
$A_{7,4,4,4}$ & 1.020 & 0.827 & 0.755 & 0.000 & \textbf{0.954} & 0.761 & 0.576 \\
$A_{7,6,6,6}$ & 0.967 & 0.881 & 0.787 & 0.000 & \textbf{0.979} & 0.788 & 0.827 \\
\bottomrule
\end{tabular}

\end{table*}

\begin{table*}
\centering

\caption{Average precision for CFF with the ``Pascal faces'' test set and different approximations relative to the exact model}
\label{tab:cff_pascal}

\begin{tabular}{c|ccccccc}
\toprule
&
Exact & ASG & PLAN & Linear I & Linear II & Quadratic I & Quadratic II
\\
\midrule
Exact & 1.000 & 0.933 & 0.774 & 0.001 & 1.039 & 0.747 & 0.893 \\
$A_1$ & 0.001 & 0.001 & 0.000 & 0.000 & 0.001 & 0.001 & 0.001 \\
$A_2$ & 0.006 & 0.006 & 0.004 & 0.000 & 0.006 & 0.003 & 0.009 \\
$A_3$ & 0.234 & 0.128 & 0.239 & 0.000 & 0.234 & 0.230 & 0.098 \\
$A_4$ & 0.357 & 0.375 & 0.259 & 0.001 & 0.361 & 0.258 & 0.370 \\
$A_5$ & 0.433 & 0.143 & 0.386 & 0.000 & 0.449 & 0.345 & 0.101 \\
$A_6$ & 0.862 & 0.844 & 0.656 & 0.000 & 0.894 & 0.638 & 0.847 \\
$A_7$ & 0.917 & 0.879 & 0.707 & 0.001 & \textbf{0.971} & 0.702 & 0.889 \\
$A_8$ & 0.725 & 0.700 & 0.520 & 0.001 & 0.697 & 0.517 & 0.617 \\
$A_{7,3,3,3}$ & 0.899 & 0.752 & 0.755 & 0.004 & 0.899 & 0.752 & 0.645 \\
$A_{7,4,4,4}$ & 0.930 & 0.889 & 0.710 & 0.000 & \textbf{0.967} & 0.708 & 0.906 \\
$A_{7,6,6,6}$ & 0.931 & 0.906 & 0.725 & 0.004 & \textbf{0.970} & 0.722 & 0.916 \\
\bottomrule
\end{tabular}

\end{table*}

Overall, the ``Linear II'' sigmoid approximation provides the best results, followed by ``ASG'', ``Quadratic I'', and ``PLAN''.
In terms of weight matrix approximations, $A_1$--$A_5$ generally give unsatisfactory results, and $A_7$ performs best.
Also, it is interesting to note that the finer approximation $A_8$ gave worse results than $A_7$.
We further evaluated some variants, where different layers of the \mbox{ConvNet} have been approximated with different sets of dyadic rationals.
For such mixed approximate structures,
we have denoted them by $A_{i,j,k,l}$,
where
the subscripts indicate the selected dyadic set for each layer.
In other words,
indices $i,j$ indicate that the dyadic sets $\mathcal{D}_i$ and $\mathcal{D}_j$,
respectively,
are employed in the first two convolution layers;
and
similarly,
indices $k,l$ correspond to the adoption
of the
dyadic sets $\mathcal{D}_k$ and $\mathcal{D}_l$,
respectively,
for
the two final fully-connected layers.
We found that the first layer requires a finer approximation than the other layers. This allowed us to maintain a good performance with very low-complexity approximations (e.g.,~$A_3$, $A_4$) for these later layers.
This can be explained according to the following:
(i)~by its own very nature, the layers have different degrees of importance;
(ii)~errors in initial layers tend to propagate through the succeeding layers;
and
(iii)~error propagation can potentially be amplified along the layers.
\begin{figure*}
  \centering
\subfigure[]{\includegraphics[width=0.52\textwidth]{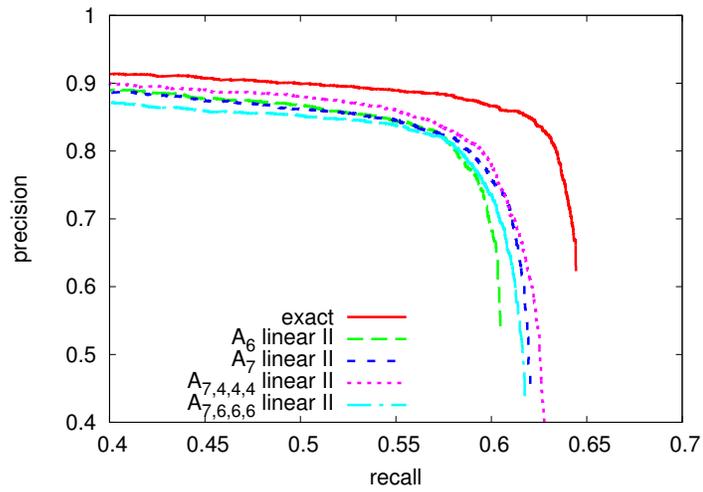}}

\subfigure[]{\includegraphics[width=0.52\textwidth]{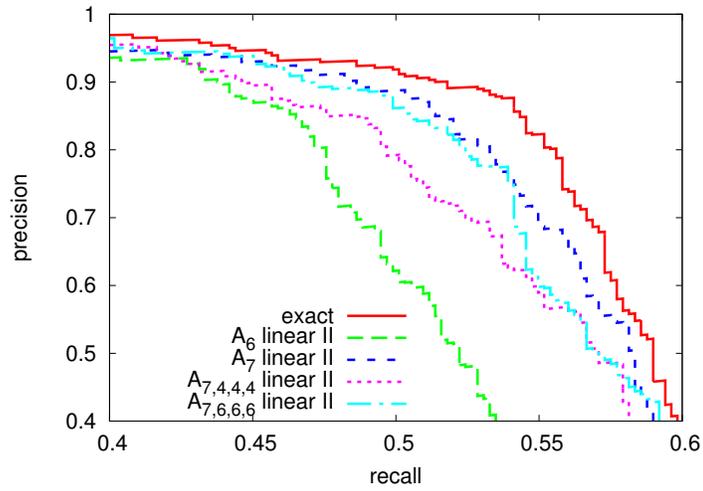}}

\subfigure[]{\includegraphics[width=0.52\textwidth]{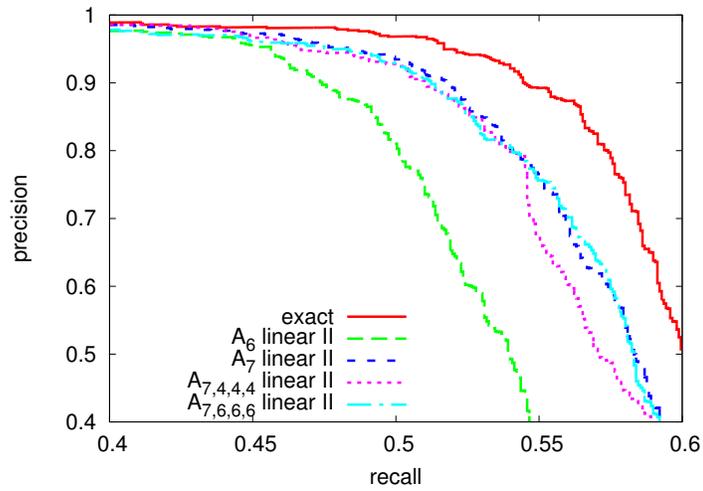}}
  \caption{%
ROC curves for the (a) FDDB, (b) AFW, and (c) Pascal datasets
comparing the
face detection performance of the original (exact) \mbox{ConvNet} model with different approximations.}
  \label{fig:cff}
\end{figure*}
Fig.~\ref{fig:cff} shows the receiver operating characteristic (ROC) curves of the best performing approximations for the three datasets.

These results are quite impressive given the fact that we considerably reduced the precision of each parameter of the \mbox{ConvNet}, and given the highly non-linear classification problem where the frontier between the face and non-face classes can be very thin and complex.

\begin{figure*}
\centering
\includegraphics[width=1.0\linewidth]{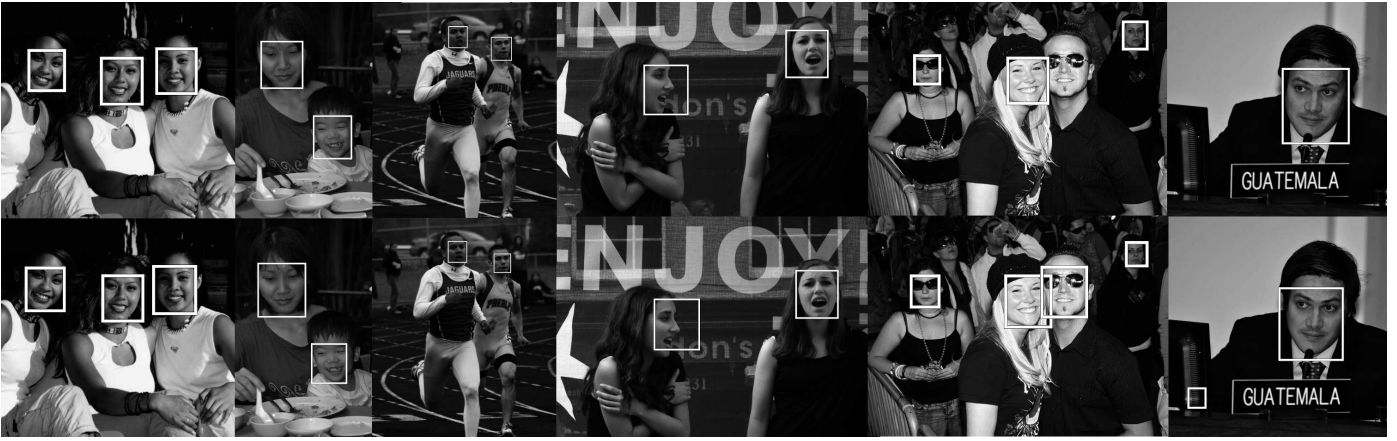}
\caption{Some CFF face detection results on the AFW dataset. \emph{Top:} exact model; \emph{bottom:} approximation $A_{7,3,3,3}$. Despite the very coarse approximation, the results are very close. In the second last image, the approximate model even detects an additional face, missed by the original CFF.\@ However, in the last example a false detection is produced.}
\label{figure-cff-results}
\end{figure*}

Fig.~\ref{figure-cff-results} shows some face detection results from the exact model (top) and the approximation $A_{7,3,3,3}$ (bottom), i.e.~a finer approximation for the first layer and a very coarse one for the rest of the layers. The results are almost identical.

\subsection{Multi-class Classification}
\label{sec:mnist}

We studied a second case where a \mbox{ConvNet} has been trained for a classical multi-class classification problem: the MNIST hand-written digit recognition dataset~\cite{lecun2015mnist}.
To show that the proposed approximations can also be applied to larger networks we trained a \mbox{ConvNet} with a different architecture containing again six layers but much more maps and around \numprint{180000} parameters and more than \numprint{5300} matrices in total.
The input is a $32\times32$ grey-scale image, and the network is composed of five convolution maps ($5\times5$ kernels) followed by five average pooling maps (connected one-to-one),
50~convolution maps ($3\times3$, fully connected),
50~average pooling maps (connected one-to-one),
100~neurons ($6\times6$ matrices, fully connected),
and
the
10~final output neurons corresponding to
the
10~digits to classify.

After having trained this \mbox{ConvNet} model on the MNIST dataset,
we approximated
all the convolution filters,
the fully-connected layer matrices
and
the activation functions which is based on the tanh-sigmoid function.
Again, all pooling coefficients and bias terms were replaced by their closest CSD representation using 8 bits.
The computation cost of the exact and approximate
structures
is shown in Table~\ref{table-cost-mnist}.
Similarly to the previous experiment,
the approximate models have totally eliminated
the multiplicative costs.
Floating-point arithmetic is not required;
being fixed-point arithmetic adequate.
The cost of the extra additions due to the CSD representation
is very low compared to the multiplicative cost
required by the exact model.
The cost of bit-shifting operations is negligible.

\begin{table*}
\centering
\caption{Arithmetic cost for MNIST-based models}
\label{table-cost-mnist}

\begin{tabular}{lcccc}
\toprule
\multirow{2}{*}{Model}
&
\multicolumn{4}{c}{Operation}
\\
\cmidrule{2-5}
&
Mult. & Add. & CSD Add. & Bit-shifting \\
\midrule
Exact &
183375 & 178110 & - & -
\\
$A_1$ &
0 & 178110 & 12740 & 23325
\\
$A_2$ &
0 & 178110 & 12722 & 23307
\\
$A_3$ &
0 & 178110 & 49127 & 59712
\\
$A_4$ &
0 & 178110 & 61228 & 71813
\\
$A_5$ &
0 & 178110 & 65211 & 75796
\\
$A_6$ &
0 & 178110 & 141401 & 151986
\\
$A_7$ &
0 & 178110 & 158595 & 169180
\\
$A_8$ &
0 & 178110 & 188417 & 199002
\\
\bottomrule
\end{tabular}
\end{table*}

\begin{figure*}
  \centering
  \includegraphics[width=0.50\textwidth]{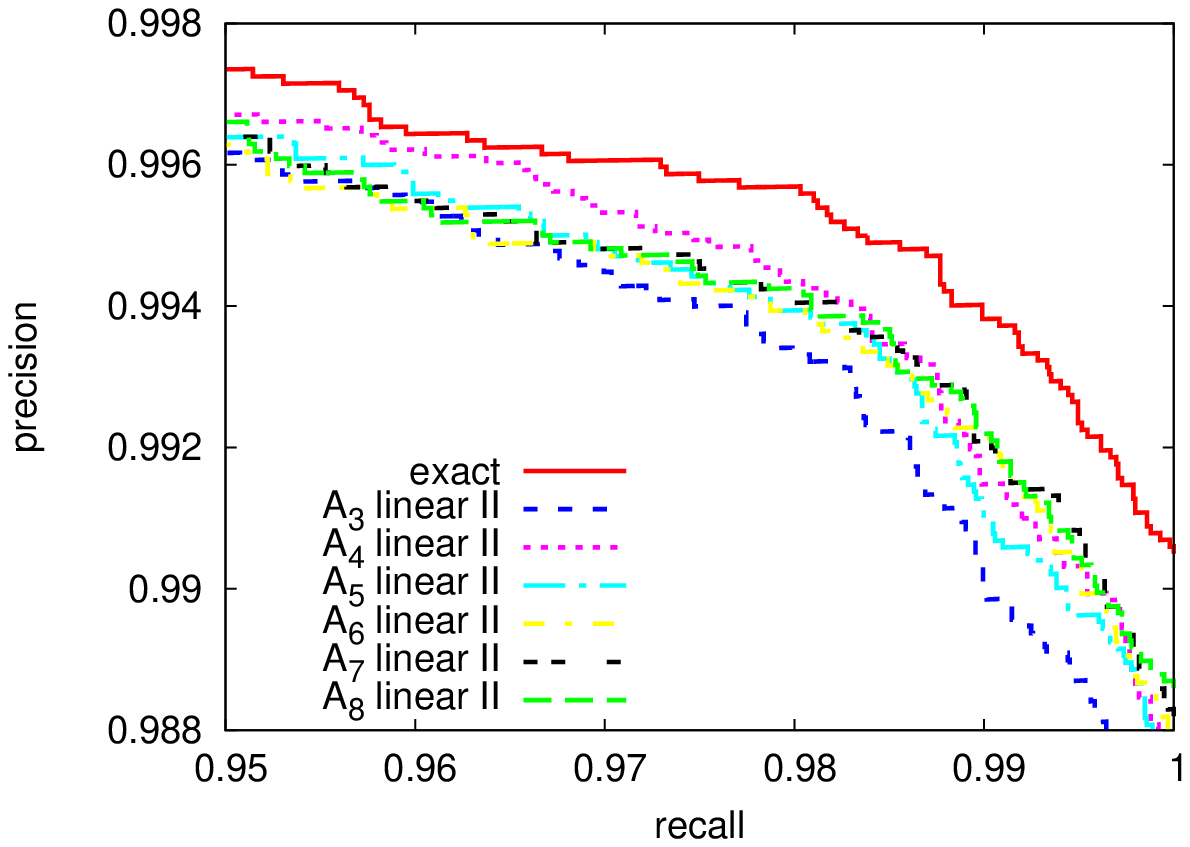}
  \caption{ROC curves for the MNIST hand-written digit classification test set comparing the performance of the original (exact) \mbox{ConvNet} model with its approximations.}
  \label{fig:mnist}
\end{figure*}

Table~\ref{tab:mnist} shows the
relative
classification rates on the MNIST test set for the different approximations, and Fig.~\ref{fig:mnist} depicts the respective ROC curves of the best-performing approximations (combined for the 10 classes).

\begin{table*}
\centering
\caption{Mean classification rates for the MNIST test set and different approximations relative to the exact model.}
\begin{tabular}{c|ccccccc}
\toprule
&
Exact & ASG & PLAN & Linear I & Linear II & Quadratic I & Quadratic II
\\
\midrule
Exact & 1.0000 & 1.0000 & 0.9847 & 0.9680 & 0.9978 & 1.0000 & 1.0000 \\
$A_1$ & 0.9684 & 0.9684 & 0.9588 & 0.9260 & 0.9615 & 0.9684 & 0.9684 \\
$A_2$ & 0.9643 & 0.9643 & 0.9627 & 0.8805 & 0.9573 & 0.9643 & 0.9643 \\
$A_3$ & 0.9961 & \textbf{0.9961} & 0.9848 & 0.9655 & \textbf{0.9944} & \textbf{0.9961} & \textbf{0.9961} \\
$A_4$ & 0.9973 & \textbf{0.9973} & 0.9863 & 0.9700 & \textbf{0.9969} & \textbf{0.9973} & \textbf{0.9973} \\
$A_5$ & 0.9976 & \textbf{0.9976} & 0.9866 & 0.9666 & \textbf{0.9969} & \textbf{0.9976} & \textbf{0.9976} \\
$A_6$ & 0.9991 & \textbf{0.9991} & 0.9868 & 0.9701 & \textbf{0.9973} & \textbf{0.9991} & \textbf{0.9991} \\
$A_7$ & 0.9992 & \textbf{0.9992} & 0.9846 & 0.9680 & \textbf{0.9977} & \textbf{0.9992} & \textbf{0.9992} \\
$A_8$ & 0.9994 & \textbf{0.9994} & 0.9848 & 0.9675 & \textbf{0.9981} & \textbf{0.9994} & \textbf{0.9994} \\
$A_{3,3,1,1}$ & 0.9931 & \textbf{0.9931} & 0.9749 & 0.9625 & \textbf{0.9924} & \textbf{0.9931} & \textbf{0.9931} \\
$A_{3,1,1,1}$ & 0.9891 & 0.9891 & 0.9684 & 0.9580 & 0.9866 & 0.9891 & 0.9891 \\
$A_{4,4,1,1}$ & 0.9937 & \textbf{0.9937} & 0.9780 & 0.9618 & \textbf{0.9943} & \textbf{0.9937} & \textbf{0.9937} \\
$A_{4,1,1,1}$ & 0.9885 & 0.9885 & 0.9655 & 0.9572 & 0.9872 & 0.9885 & 0.9885 \\
\bottomrule
\end{tabular}

\label{tab:mnist}
\end{table*}

The results show that the approximations, although very coarse, have a very small effect on the overall performance of the \mbox{ConvNet}.
In a multi-class setting, the trained \mbox{ConvNet}, at least in this particular case, is much more robust to the loss in precision of the weights induced by our approximation scheme compared to the binary classifier.
For example, a very coarse approximation like $A_{3,3,1,1}$ leads to a relative performance decrease of less than $1\%$.
\subsection{Large-Scale Deep Neural Networks}

Finally, we applied our approximation approach to a deeper and more complex network architecture, the well-known AlexNet proposed by Krizhevsky~\emph{et~al.}~\cite{krizhevskysh12}, and the ImageNet dataset~\cite{simonyanz14a} for image classification with 1000 classes.
This model contains more than $1.2$ million matrices and 5096 vectors.
We approximated all convolution filter matrices of the fully-trained 8-layer ConvNet using two different sets of dyadic rationals for different layers, a very coarse set $\mathcal{D}_9$ and a slightly finer set $\mathcal{D}_{10}$:
\begin{align*}
\mathcal{D}_9
=
&
\left\{
  -2, -1, -\frac{1}{2}, -\frac{1}{8}, 0, \frac{1}{8}, \frac{1}{2}, 1, 2
\right\}
,
\\
\mathcal{D}_{10}
=
&
\left\{
-2, -1, -\frac{1}{2}, -\frac{1}{4}, -\frac{1}{8}, 0, \frac{1}{8}, \frac{1}{4}, \frac{1}{2}, 1, 2
\right\}
.
\end{align*}
Again, all other coefficients are approximated by their closest 8-bit CSD representation.
The pooling layers do not have any coefficient here, and only linear and Rectified Linear Units (ReLU) are used as activation function, which are already of very low complexity and thus do not require any approximation.

We used the ImageNet 2012 validation set to evaluate our different approximations. And, as usual in the literature, we compute the classification accuracy as well as the top-5 accuracy for the 50000 test images.
Table~\ref{tab:imagenet} shows the results.
The approximation $A_{10}$ with the set $\mathcal{D}_{10}$ gives the best performance, with a relative decrease in accuracy of only $3.84\%$ and $2.18\%$ on the top-5 accuracy.
However, as the following line shows, we can achieve almost the same performance using the coarser set $\mathcal{D}_9$ for all convolution layers except the first one.
This again suggests that a finer approximation of the first layer is required to prevent a drastic performance drop.

\begin{table*}
\centering
\caption{Classification accuracy and top-5 accuracy for ImageNet and different approximations relative to the exact AlexNet model}
\label{tab:imagenet}
\begin{tabular}{c|cccc}
\toprule
& \multicolumn{2}{c}{Absolute} & \multicolumn{2}{c}{Relative} \\
        & Accuracy & Top-5 & Accuracy & Top-5 \\
\midrule
Exact    & 0.5682 & 0.7995 & 1.0000 & 1.0000 \\
$A_9$    & 0.4862 & 0.7288 & 0.8558 & 0.9117 \\
$A_{10}$ & \textbf{0.5463} & \textbf{0.7820} & \textbf{0.9616} & \textbf{0.9782} \\
$A_{10,9,9,9,9,9}$ & 0.5423 & 0.7794 & 0.9544 & 0.9750 \\
$A_{10,10,9,9,9,9}$ & 0.5442 & 0.7796 & 0.9578 & 0.9751 \\
\bottomrule
\end{tabular}
\end{table*}

\section{Conclusion}
\label{section-conclusion}

We presented a novel scheme for approximating the parameters of a trained \mbox{ConvNet}, notably the convolution filters, neuron weights, as well as pooling and bias coefficients.
Activation functions
were
also approximated.
The particularity of the matrix approximations is that they allow for an extremely efficient implementation---software or hardware---using only additions and bit-shifts, and no multiplication.
We thoroughly evaluated the impact of this parameter approximation measuring the overall performance of \mbox{ConvNet}s on
three
different use cases:
one smaller \mbox{ConvNet} for face detection,
a larger \mbox{ConvNet} for hand-written digit classification,
and
a much more complex, deep \mbox{ConvNet} for large-scale image classification.

For all three models, our proposed scheme was able to produce low-complexity approximations without a significant loss in performance.

These results suggest that huge reductions in computational complexity of trained \mbox{ConvNet} models can be obtained, and extremely efficient hardware implementations can be realized.
Further studies need to be undertaken to analyse the impact of this type of approximations for more use cases and different architectures.

{\small
\singlespacing
\bibliographystyle{siam}
\bibliography{final-ref}
}

\end{document}